\documentclass{article}
\usepackage[preprint]{neurips_2024}
\usepackage[utf8]{inputenc} 
\usepackage[T1]{fontenc}    
\usepackage{hyperref}       
\usepackage{url}            
\usepackage{booktabs}       
\usepackage{amsfonts}       
\usepackage{nicefrac}       
\usepackage{microtype}      
\usepackage{xcolor}         
\usepackage{xspace}
\usepackage{header}
\usepackage{multirow}
\usepackage{colortbl}
\usepackage{graphicx}
\usepackage{enumerate}
\usepackage{tikz}
\usepackage{CJK}
\usepackage{colortbl}
\usepackage[most]{tcolorbox}
\newtcolorbox[auto counter, number freestyle={\noexpand\arabic{\tcbcounter}}]{definedbox}[2][]{%
    enhanced,
    colback=black!5!white,
    colframe=black!75!white,
    title=Example~\thetcbcounter: #2,
    #1
}
\newtcolorbox[auto counter, number freestyle={\noexpand\arabic{\tcbcounter}}]{promptbox}[2][]{%
    enhanced,
    colback=teal!5!white,
    colframe=teal!50!black,
    title=Template~\thetcbcounter: #2,
    #1
}

\title{\algo: A Chinese Legal Knowledge-Enhanced Large Language Model}

\author{
  Zhi Zhou$^{1 \dagger}$, Jiang-Xin Shi$^{1 2 \dagger}$, Peng-Xiao Song$^{1 \dagger}$, Xiao-Wen Yang$^{1 2 \dagger}$,\\
  \textbf{Yi-Xuan Jin$^{1}$, Lan-Zhe Guo$^{1 3 \ddagger}$, Yu-Feng Li$^{1 2 \ddagger}$}\\
  $^1${National Key Laboratory for Novel Software Technology, Nanjing University}\\
  $^2${School of Artifical Intelligence, Nanjing University}\\
  $^3${School of Intelligence Science and Technology, Nanjing University}\\
  \texttt{\{zhouz,shijx,songpx,yangxw,jinyx,guolz,liyf\}@lamda.nju.edu.cn} \\
  \hspace{-0.25cm}$^{\dagger}$ Equal Contribution  \ \ $^{\ddagger}$ Corresponding Author
}

\begin{document}

\maketitle

\begin{abstract}
  Large language models (LLMs), including both proprietary and open-source models, have showcased remarkable capabilities in addressing a wide range of downstream tasks. 
  Nonetheless, when it comes to practical Chinese legal tasks, these models fail to meet the actual requirements. 
  Proprietary models do not ensure data privacy for sensitive legal cases, while open-source models demonstrate unsatisfactory performance due to their lack of legal knowledge. 
  To address this problem, we introduce \algo, the first open-source model specifically designed for Chinese legal applications. 
  \algo comprises two key components: legal-oriented pre-training and legal supervised fine-tuning. 
  Specifically, we employ large-scale Chinese legal documents for legal-oriented pre-training to incorporate legal domain knowledge. 
  To further improve the model's performance on downstream legal tasks, we create a knowledge-driven instruction dataset for legal supervised fine-tuning. 
  Our experimental results demonstrate that \algo outperforms the open-source LLaMA 7B model. 
  Our code and resources are publicly available at \url{https://github.com/pengxiao-song/LaWGPT} and have received 5.7K stars on GitHub.
\end{abstract}

\section{Introduction}
\label{sec:introduction}

Large language models (LLMs)~\citep{GPT4,hugo23LLaMA2} have achieved remarkable success in various natural language processing (NLP) tasks, including natural language understanding~\citep{dong19understanding}, reasoning~\citep{DBLP:conf/acl/0009C23}, and generation~\citep{yu22text}. 
Both proprietary and open-source LLMs exhibit strong generalization capabilities, enabling their application in diverse downstream scenarios, such as medicine~\citep{th23medicine}, finance~\citep{DBLP:journals/corr/abs-2306-06031}, education~\citep{DBLP:conf/bigdataconf/GanQWL23}. 
Recent studies~\citep{li2023LawBench,nguyen2023LawGPT} have demonstrated the preliminary effectiveness of existing general LLMs in legal tasks, including legal judgment prediction~\citep{luo2017legalpredict}, legal documents retrieval~\citep{chen2013textmining}, and legal question answering~\citep{zhong2020nlpbenefit}. 

Despite the preliminary effectiveness of LLMs in legal applications, there are two obstacles that hinder their practical use in legal tasks. 
On the one hand, proprietary LLMs such as GPT-4~\citep{GPT4} and GPT-3.5 Turbo~\citep{GPT35T} can only be accessed through APIs, which do not guarantee data privacy in sensitive legal cases. 
On the other hand, open-source LLMs like LLaMA~\citep{hugo23LLaMA} and ChatGLM~\citep{du2022glm} fail to achieve satisfactory performance due to their insufficient legal knowledge and incompatibility with downstream legal tasks. 
Therefore, it is necessary to develop a open-source LLM specifically designed for legal applications  in order to overcome the existing obstacles. 

In this paper, we introduce \algo, the first open-source Chinese legal knowledge-enhanced large language model. With the advantage of being open-source, \algo can be self-hosted and accessed privately to ensure data privacy, as compared to proprietary models. We then present legal-oriented pre-training, which utilizes our large-scale legal pre-training corpus to incorporate domain-specific legal knowledge into \algo, improving its understanding, reasoning, and generation foundational capabilities in legal tasks. 
Additionally, we propose legal supervised fine-tuning, employing our knowledge-driven instruction dataset to further enhance \algo's performance on downstream legal tasks. 
Experimental results demonstrate that \algo surpasses the open-source LLaMA 7B model in major legal tasks, shedding light on the development of a practical Chinese legal LLM. 

In summary, our contributions can be summarized as follows: 
\begin{enumerate}[(a)]
    \item We present the first open-source Chinese legal knowledge-enhanced large language model \algo. The code and model are available on GitHub~\footnote{\url{https://github.com/pengxiao-song/LaWGPT}} and have received 5.7K stars.
    \item We construct a comprehensive legal pre-training corpus and propose a legal-oriented pre-training approach to enhance \algo's foundational abilities in legal tasks by integrating domain-specific knowledge.
    \item We create a knowledge-driven instruction dataset and utilize legal supervised fine-tuning to further adapt \algo to various legal tasks and improve its downstream performance.
    \item Our experimental results demonstrate that \algo achieves better performance than the open-source LLaMA 7B model across major legal tasks, providing strong evidence for the effectiveness of our proposed model.
\end{enumerate}

\section{Related Work}
\label{sec:relatedwork}

In this section, we review the existing work on addressing legal tasks using LLMs. This focus is on general language models, legal language models, and legal benchmarks as follows. 

\subsection{General Language Models}

Benefiting from training with large scale corpus, recent LLMs have shown impressive performance on various kind of downstream tasks, including legal tasks. Recent LLMs, trained on extensive corpora, have demonstrated impressive performance across a variety of downstream tasks, including tasks in the legal domain. 
Proprietary LLMs, such as GPT-4~\citep{GPT4}, GPT-3.5-Turbo~\citep{GPT35T}, PaLM~\citep{Chowdhery23PaLM}, and PaLM2~\citep{Rohan23PaLM2}, exhibit strong capabilities in handling legal tasks. 
Their impressive performance not only demonstrates the potential of LLMs in addressing legal tasks but also facilitates the low-cost, automated construction of high-quality datasets.
Concurrently, open-source LLMs, such as LLaMA~\citep{hugo23LLaMA}, LLaMA2~\citep{hugo23LLaMA2}, MPT~\citep{MosaicML2023mtp7B}, ChatGLM 2~\citep{du2022glm}, and Baichuan 2~\citep{aiyun23baichuan}, are available in various model scales. These open-source models facilitate the fine-tuning of legal-specific models using targeted legal datasets, potentially enhancing performance.

\subsection{Legal Language Models}

The legal language models are fine-tuning based on pre-trained language models or trained from scratch using legal related data for improving legal capabilities of models. 
Early research in this field utilized a model architecture with millions of parameters for various legal tasks. 
Extensive efforts have been made to address these tasks separately. These include legal judgment prediction~\citep{luo2017legalpredict, chalkidis19legalpredict, yang2019legalpredict}, legal documents and cases retrieval~\citep{chen2013textmining, shao2020bertpli, li2023SAILER}, legal reading comprehension~\citep{duan2019cjrc}, and legal question answering~\citep{zhong2020nlpbenefit, phi2020legalqa}. 
With the benefit of pre-trained models~\citep{chalkidis2020legalbert, cui21roberta}, Lawformer~\citep{xiao21lawformer} combines three attention mechanisms to address the problem of long legal documents, covering a variety of legal tasks. 
Recent advances of LLMs have given rise to legal LLMs work. 
HanFei~\citep{zhang2023HanFei}, LexiLaw~\citep{li2023SAILER}, LawGPT-zh~\citep{liu2023lawgptzh}, and LawGPT-1.0~\citep{nguyen2023LawGPT}  fine-tune foundational LLMs using a specially constructed or collected legal corpus to enhance their legal capabilities. 
To tackle the hallucination problem in legal tasks, LLMs such as ChatLaw~\citep{cui23chatlaw}, Wisdom-Interrogatory~\citep{wu2024WisdomInterrogatory}, and Lawyer-LLaMA~\citep{huang2023LawyerLLaMA} incorporate a legal data retrieval method to improve the robustness of their responses.
LawGiBa~\citep{nguyen2023LawGiBa}, based on the GPT-4 model, has established a legal system.
Fuzi-Mingcha~\citep{deng2023fuzi} has created a legal syllogistic reasoning dataset for fine-tuning to ensure logical format and accurate reasoning results.

\subsection{Legal Benchmarks}

With the emergence of enormous language models for legal tasks, several benchmarks have been proposed to evaluate a variety of existing models.
LawBench~\citep{li2023LawBench} collects 20 legal tasks within three cognitive levels, i.e., legal knowledge memorization, understanding, and applying, to thoroughly evaluate the performance of existing models. LAiW~\citep{dai2023LAiw} contains 14 tasks to evaluate the legal capabilities of LLMs from three levels, i.e., basic information retrieval, legal foundation inference, and complex legal application.
SimuCourt~\citep{he2024simucourt} introduces a judicial decision-making task to evaluate the judicial analysis and decision-making power of LLMs.

\section{Methodology}
\label{sec:methodology}

\begin{figure}[t]
    \begin{definedbox}[label=ex:pre-training-corpus]{Legal Pre-training Corpus}
        \begin{CJK*}{UTF8}{gbsn}
        上诉人*********（以下简称**学校）因与被上诉人************（以下简称**公司）装饰装修合同纠纷一案，不服*********人民法院（20xx）辽****民初****号民事判决，向本院提起上诉。本院依法组成合议庭审理了本案。本院认为 本院认为，一审判决程序违法。
        1.**学校一审反诉请求解除装修合同及空调合同，一审仅判决解除装修合同，空调合同应否解除未予审理，属漏审漏判；一审双方当事人均未提出解除案涉补充协议，一审判决解除补充协议，超出当事人的诉请；
        2.**学校一审反诉请求要求**公司按已付工程款数额开具发票，一审仅判决**公司给付欠付款项的发票，亦属漏审漏判；
        3.**公司起诉状中明确了监控布线款为15600元，在**公司未提出变更诉讼请求的情况下，一审按照鉴定结论中的数额认定监控布线损失为32868.85元，亦属超出当事人的诉请。 
        一审判决认定事实不清。案涉工程两个施工合同均为固定总价合同，一审判决**公司给付**学校欠付工程款258449.56元依据的鉴定结论计算方式错误，且装修合同和空调合同应分别计算。另外，案涉已完工程造价鉴定中是否包含了已施工与图纸不符、质量不合格修复部位的工程造价？已施工与图纸不符、质量不合格修复费用的鉴定结论是否为修复到施工前的原始状态？重审时需补充鉴定。 一审重审时应围绕双方当事人的诉讼请求，合理分配举证责任，在查清事实的基础上依法裁判。 综上，依照《中华人民共和国民事诉讼法》第一百七十七条第一款第（三）、（四）项之规定，裁定如下：判决结果 一、撤销*********人民法院作出的（20xx）辽****民初****号民事判决； 二、本案发回*********人民法院重审。 上诉人*********学校预交的二审案件受理费6579元予以退回。\end{CJK*}
    \end{definedbox}
    \vskip -0.1in
\end{figure}

\begin{figure}[t]
    \begin{definedbox}[label=ex:sft-corpus-a]{Legal-supervised Dataset - Subset (a)}
        \begin{CJK*}{UTF8}{gbsn}
        \textbf{Instruction:} 请问我向借钱人要钱多次未果，向法院起诉，法院多久才立案
        \tcbline
        \textbf{Output:} 起诉的当日 ，法院就会立案的。\end{CJK*}
    \end{definedbox}
    \vskip -0.1in
\end{figure}

\begin{figure}[t]
    \begin{definedbox}[label=ex:sft-corpus-b]{Legal-supervised Dataset - Subset (b)}
        \begin{CJK*}{UTF8}{gbsn}
        \textbf{Instruction:} 根据《中华人民共和国海商法》，在海事关系的法律适用中，旗国法适用于下列哪些情形? (A) 船舶抵押权的设定 (B) 同国籍船舶在公海发生碰撞的损害赔偿 (C) 共同海损理算 (D) 海事赔偿责任限制.
        \tcbline
        \textbf{Output:} (A) 船舶抵押权的设定(B) 同国籍船舶在公海发生碰撞的损害赔偿
        \end{CJK*}
    \end{definedbox}
    \vskip -0.1in
\end{figure}

\begin{figure}[t]
    \begin{promptbox}[label=ex:sft-corpus-c]{Prompt of ChatGPT for Augmentation}
        \begin{CJK*}{UTF8}{gbsn}
        我希望你担任语言专家的角色。我会给你一段与法律问答文本，请你使用正式的文风润色它。要求：$\backslash$n
            1. 修正语法错误、标点符号错误，去掉特殊符号，必须使语句更通顺。
            2. 使逻辑更清晰、格式更规范，比如向<answer>中换行符。
            3. 使更礼貌，比如向<question>中加入“请问”等礼貌用语。
            4. 不要写任何解释性语句。
            5. <question>应该是问题，<answer>应该是答案。
            这段对话是：$\backslash$n<question>:\{instruction\} $\backslash$n<answer>:\{output\} $\backslash$n$\backslash$n
        以JSON格式返回结果：
        \end{CJK*}
    \end{promptbox}
    \vskip -0.1in
\end{figure}

\begin{figure}[t]
    \begin{promptbox}[label=ex:alpaca]{Alpaca Training Template}
        Below is an instruction that describes a task. Write a response that appropriately completes the request.$\backslash$n$\backslash$n \#\#\# Instruction:$\backslash$n\{instruction\}$\backslash$n$\backslash$n \#\#\# Response: $\backslash$n\{output\}
    \end{promptbox}
    \vskip -0.1in
\end{figure}

\begin{figure}[t]
    \begin{promptbox}[label=ex:alpaca-test]{Alpaca Testing Template}
        Below is an instruction that describes a task. Write a response that appropriately completes the request.$\backslash$n$\backslash$n \#\#\# Instruction:$\backslash$n\{instruction\}$\backslash$n$\backslash$n \#\#\# Response: $\backslash$n
    \end{promptbox}
    \vskip -0.1in
\end{figure}

In this section, we introduce our \algo, a large language model specifically designed for Chinese legal applications, aimed at effectively addressing various downstream legal tasks. 
\algo addresses the two major challenges in applying existing open-source general LLMs to legal tasks:
\begin{enumerate}[(a)]
    \item The lack of legal domain knowledge in open-source general LLMs, which is crucial for performing legal tasks effectively;
    \item The insufficient training of open-source general LLMs on downstream legal tasks, resulting in suboptimal performance in legal applications.
\end{enumerate}
We apply legal-oriented pre-training to \algo to incorporate legal domain knowledge within the open-source base model. 
Then, we conduct legal supervised fine-tuning to further enhance \algo's performance on downstream legal tasks.
Each component is elaborated as follows. 

\subsection{Legal-Oriented Pre-Training}

General LLMs are typically pre-trained on large-scale general corpus, which may lack sufficient legal domain knowledge. Consequently, this can result in a limited understanding and reasoning ability for legal tasks. To address this limitation, we propose the integration of \textbf{L}egal-oriented \textbf{P}re-\textbf{T}raining (\text{LPT})  into \algo, aiming to enhance its legal domain knowledge.

To incorporate legal domain knowledge into \algo, we collect a large-scale legal pre-training corpus $\mathcal{D}_{\text{LPT}}$ consisting of 500K legal documents from various legal domains, including civil law, criminal law, and administrative law. 
Example~\ref{ex:pre-training-corpus} presents a civil-law legal document from the legal pre-training corpus. For each legal document, the tokenizer of base model encodes the text into a token sequence $\boldsymbol{x}=(x_0, x_1, \ldots)$, and we perform legal-oriented pre-training on the base model $f_{\Theta}(\cdot)$ in an autoregressive manner using the following objective:
\begin{equation}
\mathcal{L}_{\text{LPT}}(\Theta, \mathcal{D}_{\text{LPT}}) = \mathbb{E}_{\boldsymbol{x} \sim \mathcal{D}_{\text{LPT}}}\left[ -\sum\limits_{i} \log{f_{\Theta}(x_i|x_0, x_1, \ldots, x_{i-1})}\right]
\end{equation}
where $x_0, x_1, \ldots, x_{i-1}$ denote the context tokens, $x_i$ denotes the target token, and $\Theta$ is the parameters of base model $f_{\Theta}(\cdot)$. 
We optimize the parameters of base model $\Theta$ using $\mathcal{L}_{\text{LPT}}$ to obtain the parameters of legal-oriented pre-trained model $\Theta^{\text{LPT}}$.

\subsection{Legal-Supervised Fine-Tuning}

Although $f_{\Theta^{\text{LPT}}}(\cdot)$ has been pretrained with legal domain knowledge, it is not optimal for specific downstream legal tasks as it cannot generate the desired responses by following the instructions. 
To address this issue, we propose \textbf{L}egal-{S}upervised \textbf{F}ine-\textbf{T}uning (\text{LFT}) to further adapt \algo to various downstream legal tasks. 
Specifically, we construct a 300K knowledge-driven instruction dataset, $\mathcal{D}_{\text{LFT}}$, consisting of three subsets: 
\begin{enumerate}[(a)]
    \item An open-source dataset~\footnote{https://github.com/liuhuanyong/CrimeKgAssistant} with 200K samples, which includes crime type prediction and crime consult tasks to fine-tune the model for better understanding of crime-related legal tasks and generating user-friendly responses; 
    \item The JEC-QA dataset~\citep{zhong20jecqa} with 20K samples, which consists of legal question answering tasks to fine-tune the model for better adaptation to legal downstream tasks;
    \item A constructed legal datasets with 80K samples by refining subsets (a) and (b) with ChatGPT~\citep{GPT35T}, which augments more high-quality legal QA samples, thereby enhancing the generalizability of the model.
\end{enumerate}
The subsets (a) and (b) are shown in Examples~\ref{ex:sft-corpus-a} and~\ref{ex:sft-corpus-b}, respectively.  The subset (c) is reinfed using the prompt template in Template~\ref{ex:sft-corpus-c} to augment the samples in subsets (a) and (b), where we replace <instruction> with real questions and <output> with the corresponding answer. 
We adopt the Stanford Alpaca template~\citep{taori23alpaca} in Template~\ref{ex:alpaca} to wrap the instruction and output in our dataset.
Then, the parameters of our pre-trained model $\Theta^{\text{LPT}}$ are fine-tuned on $\mathcal{D}_{\text{LFT}}$ using the following objective:
\begin{equation}
    \mathcal{L}_{\text{LFT}}(\Theta, \mathcal{D}_{\text{LFT}}) = \mathbb{E}_{\boldsymbol{x} \sim \mathcal{D}_{\text{LFT}}}\left[ -\sum\limits_{i\in\{\text{output}\}} \log{f_{\Theta^{\text{LPT}}}(x_i | x_0, x_1, \ldots, x_{i-1})} \right]
\end{equation}
where $\Theta$ represents the optimized parameters, $\boldsymbol{x} = (x_0, x_1, \ldots)$ represents the tokenized input sequence drawn from dataset $\mathcal{D}_{\text{LFT}}$ and wrapped by Template~\ref{ex:alpaca}, and $\{\text{output}\}$ represents the index set of the output tokens. 
We optimize the our pre-trained parameters $\Theta^{\text{LPT}}$ to obtain the parameters of \algo $\Theta^{\text{LFT}}$.

\subsection{Inference of \algo}

When applying \algo to downstream tasks, we should wrapped the instruction using the Alpaca template in Template~\ref{ex:alpaca-test} and then tokenized the texts into $\boldsymbol{x} = (x_0, x_1, \ldots, x_n)$. 
Then, we feed the tokenized input sequence $\boldsymbol{x}$ into the fine-tuned model $f_{\Theta^{\text{LFT}}}(\cdot)$ to generate the response in an autoregressive manner.

\section{Experiments}
\label{sec:experiments}

\subsection{Implementation Details}

We trained \algo using 8 NVIDIA V100 GPUs, based on the Chinese-Alpaca-Plus 7B base model~\citep{chinese-llama-alpaca}, in two stages: legal-oriented pre-training, and legal-supervised fine-tuning.
For legal-oriented pre-training, we adopt our 500K legal pre-training corpus $\mathcal{D}_{\text{LPT}}$ to train the base model using the LoRA technique~\citep{hu22lora}. We set the LoRA rank to 16, alpha to 32, and dropout to 0.05. The learning rate was set to 0.0003, the batch size to 128, and the training epoch to 1. 
For legal-supervised fine-tuning, we adopt our 30K legal-supervised corpus $\mathcal{D}_{\text{LFT}}$ to fine-tune our pre-trained model with Alpaca template using the LoRA technique. We set the LoRA rank to 8, alpha to 16, and dropout to 0.05. We set the learning rate to 0.0003, the batch size to 64, and the training epoch to 20.

\subsection{Performance Evaluation}

\begin{table}[th]
    \caption{Performance comparison between \algo, proprietary models including GPT-3.5 Turbo~\citep{GPT35T} and GPT-4~\citep{GPT4}, and 7B open-source model LLaMA~\citep{hugo23LLaMA} on the zero-shot setting. The best performance among \algo and open-source models is in bold.}
    \label{tab:zeroshot}
    \centering
    \begin{tabular}{l|rrrrrrrrrrrrrrrrrr|r}
        \toprule
        \multirow[l]{2}{*}{Models} & \multicolumn{9}{c}{Tasks} \\ \cmidrule{2-10}
        & \#1 & \#2 & \#3 & \#4 & \#5 & \#6 & \#7 & \#8 & Avg. \\
        \midrule
        \rowcolor[gray]{.90} GPT-3.5 Turbo & 29.5 & 31.3 & 35.5 & 78.7 & 76.8 & 27.4 & 61.2 & 17.4 & 44.7 \\
        \rowcolor[gray]{.90} GPT-4 & 52.5 & 27.5 & 42.0 & 82.6 & 81.9 & 48.6 & 77.6 & 19.6 & 54.0 \\
        \midrule
        LLaMA  & \textbf{1.0} & 7.5 & 7.0 & 41.3 & \textbf{54.2} & 0.2 & 14.4 & \textbf{7.8} & 16.7 \\
        LaWGPT & 0.2 & \textbf{11.0} & \textbf{15.7} & \textbf{42.4} & 40.8 & \textbf{6.2} & \textbf{15.4} & 7.6 & \textbf{17.4} \\
        \bottomrule
    \end{tabular}
\end{table}

In this section, we conduct experiments to evaluate the performance of \algo on 8 legal applications~\citep{li2023LawBench}, including fact-based article prediction (\#1), scene-based article prediction (\#2), charge prediction (\#3), prison term prediction without article (\#4), prison term prediction with article (\#5), case analysis (\#6), criminal damages calculation (\#7), and consultation (\#8), in a zero-shot setting. 
We compare the performance of \algo with proprietary models including GPT-3.5 Turbo~\citep{GPT35T} and GPT-4~\citep{GPT4}, and 7B open-source models including LLaMA~\citep{hugo23LLaMA}. The results are shown in Table~\ref{tab:zeroshot}. 
The results show that our \algo outperforms LLaMA 7B model on major tasks and leading to a better average performance. 
Despite the advantage of preserving data privacy, there is still a significant performance gap between \algo and proprietary models. This result inspires us and the following researchers to explore the potential of \algo in the future work. 

\section{Conclusion}
\label{sec:conclusion}

In this technical report, we introduce \algo, a Chinese legal knowledge-enhanced large language model specifically designed for Chinese legal applications. We introduce the legal-oriented pre-training and legal supervised fine-tuning to incorporate legal domain knowledge and enhance the model's performance on downstream legal tasks, respectively. Our experimental results demonstrate that \algo outperforms the open-source LLaMA 7B model.  
We hope this technical report and \algo model can inspire future research on Chinese legal applications and contribute to the development of the legal AI community. 

\newpage
{
\small
\bibliographystyle{plainnat}
\bibliography{ref}
}

\end{document}